\documentclass{article}

\usepackage{threeparttable}
\usepackage{arxiv}
\usepackage[utf8]{inputenc} 
\usepackage[T1]{fontenc}    
\usepackage{hyperref}       
\usepackage{url}            
\usepackage{booktabs}       
\usepackage{amsfonts}       
\usepackage{nicefrac}       
\usepackage{microtype}      
\usepackage{lipsum}
\usepackage{graphicx}
\usepackage[super]{nth}
\usepackage{xspace}
\usepackage{diagbox}

\def\etc{\emph{etc.}\xspace} 
 
\newcommand{\etal}{\textit{et al.}}

\usepackage{cite}

\title{Ultra Light OCR Competition Technical Report}

\author{
 Shuhan Zhang, Yuxin Zou, Tianhe Wang, Yichao Xiong \\
  \texttt{\{oozorakuro409, zyx.dongtai, wth625361486, jonathan.odd\}@gmail.com}
}

\begin{document}
\maketitle

\section{Introduction}

\par Ultra Light OCR Competition is a Chinese scene text recognition competition jointly organized by \textbf{CSIG} (China Society of Image and Graphics) and \textbf{Baidu, Inc}. In addition to focusing on common problems in Chinese scene text recognition, such as long text length and massive characters, we need to balance the trade-off of model scale and accuracy since the model size limitation in the competition is 10M. 

\par From experiments in aspects of data, model, training, \etc, we proposed a general and effective method for Chinese scene text recognition, which got us \textbf{\nth{2}} place among over 100 teams with accuracy \textbf{0.817} in TestB dataset. The code is available at \url{https://aistudio.baidu.com/aistudio/projectdetail/2159102}.

\section{Method}

\par With limited resources(on average, we only had access to 4 NVIDIA 1080Ti graphical cards during this competition) and limited time, we decided to focus on one goal at a time. We start by trying to push up the accuracy, then as the competition moving to the end, we shrink the model with the least accuracy drop we can achieve. Our method can be divided into two parts: Data Analysis and Actual Experiments. We divide our experiments into 3 stages: Over-fitting\&baseline, data, model. 

\subsection{Insights from Data}

\par The dataset provided includes both street-view images(such as shop signs, landmarks, \etc) and images taken from documents and websites. All images provided are rectified to avoid irregular text problems which is a common problem in scene text recognition.

\par There are a total of 120,000 images provided, of which 100,000 images (with annotations) forms the train set, and the rest are divided into two test sets (TestA/TestB), each with 10,000 images (without annotations). 

\par We randomly divide the train set provided into train/val set with ratio 9:1 for the convenience of subsequent experiments and analysis since online submissions for TestA are limited to 5 times a day.

\subsubsection{Annotation}
\paragraph{Distinct characters} The numbers of distinct characters in the train set and val set were shown in Table \ref{tab:amount}. As anticipated, the number of possible characters is enormous. Note that there are a total of 66 characters(from 63 text samples) that only appear in the val set, which would induce 0.63\% accuracy loss theoretically.

\begin{table}[htb]
    \centering
    \begin{tabular}{ccc}
    \hline
         & Number of Distinct Characters\\
    \hline
         Train Set  & 3908\\
         Val Set & 2516\\
    \hline
        Total & 3974\\
    \hline
    \\
    \end{tabular}
    \caption{Number of Distinct Characters across Train/Val Set}
    \label{tab:amount}
\end{table}

\paragraph{Character Frequency} As shown in Table \ref{tab:frequency}, we found that, in terms of character frequency, the long-tail problem is significant. Note that around 1/5 characters appeared only once.
\begin{table}[htb]
    \centering
    \begin{tabular}{ccccccc}
    \hline
            &1000+ &100-1000  &10-100    &1-10  &1     &Total\\
    \hline
    Train   &86(2.2\%)&783(20.0\%)&1246(31.9\%)&1138(29.1\%)&655(16.8\%)&3908\\
    Val     &3(0.1\%)&99(3.9\%)&812(32.3\%)&1037(41.2\%)&565(22.5\%)&2516\\
    \hline
    Total   &100   &828       &1262      &1116  &668   &6424\\
    \hline
    \\
    \end{tabular}
    \caption{Character Frequency in Train/Val Set}
    \label{tab:frequency}
\end{table}

\begin{figure}[htb]
    \centering
    \includegraphics[scale=0.7]{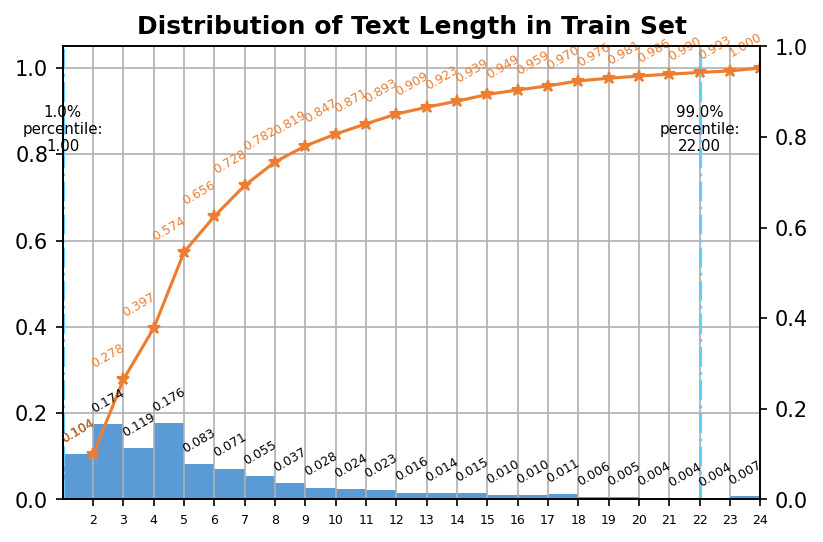}
    \caption{Distribution of Text Length in Train Set}
    \label{fig:text_length}
\end{figure}

\paragraph{Text Length} As shown in Figure \ref{fig:text_length}, most of text samples have length in range 1 to 22. Here we only show the text length distribution for train set since val set has similar distribution as train set.

\paragraph{Label Noise} We random sampled and eyeballed 500 images from train set and noticed that 17 of them were mislabeled. This means that there are noises in train/val set and potentially in test set. As a result, theoretically, assuming there are noises of same level in the test set, a perfect model would suffer a 3.4\% accuracy loss.

\subsubsection{Image} 
\begin{figure}[htb]
    \centering
    \includegraphics[scale=0.6]{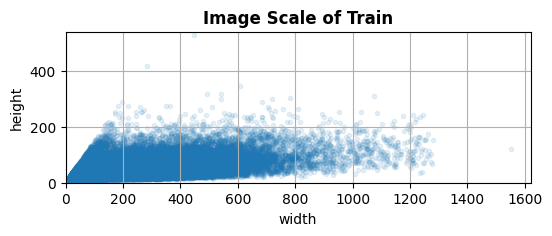}
    \caption{Image Scale Distribution of Train Set}
    \label{fig:train_image_scale}
\end{figure}

\paragraph{Image Scale} The scale of images in train set is shown in Figure \ref{fig:train_image_scale}. As expected, most images have longer width than height. We also notice that the aspect ratio of the dataset is limited in certain range since all the image in the dataset form two clear slop in the figure.

\begin{figure}[htb]
    \centering
    \includegraphics[scale=0.4]{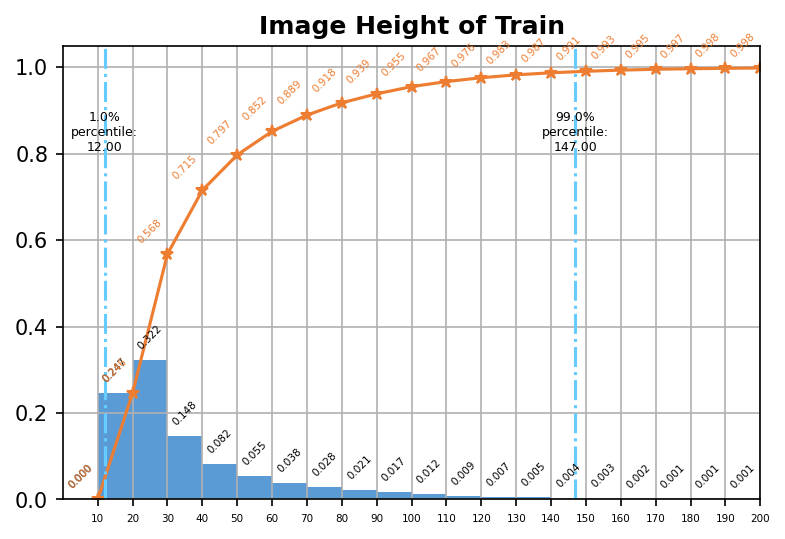}
    \caption{Image Height in Train Set}
    \label{fig:train_image_height}
\end{figure}

\paragraph{Image Height} As for the height, as showed in Figure \ref{fig:train_image_height}, the height of 39.4\% of the images are greater than 32 pixels, 21.6\% are greater than 48 pixels, and 13.1\% are greater than 64 pixels.

\subsection{Experiment stages}
\subsubsection{Baseline}

\par We called this stage ‘over-fitting\&baseline’ earlier because we'd like to emphasize the importance of fitting the train set. Here, over-fitting refers to a set of experiments that tries to ‘over-fit’ the model on the train set. These experiments are conducted on different subsets of training set: starting from a single image to a batch of images, to a larger portion of the train set, and eventually the whole train set. The goal of those experiments is to spot problems in the early stage. For instance, if a model cannot even over-fit on only one batch of images, it's obvious that there is something wrong in the training process: not enough model capacity? Inappropriate hyper-parameters? Some special cases of data that you are not aware of? 
\par We start over-fitting experiments with settings below:

\paragraph{Training Framework} According to the competition regulations, our only choice is PaddleOCR\footnote{\url{https://github.com/PaddlePaddle/PaddleOCR}}. 

\paragraph{Algorithm} Although the attention-based prediction approaches have become the mainstream method in the field of scene text recognition, we choose CTC-based method after survey investigation because of its strength on dealing with long text sequences and large-scale category recognition tasks\cite{DBLP:journals/corr/abs-2005-03492}. 

\par
\begin{figure}[htb]
    \centering
    \includegraphics[scale=0.5]{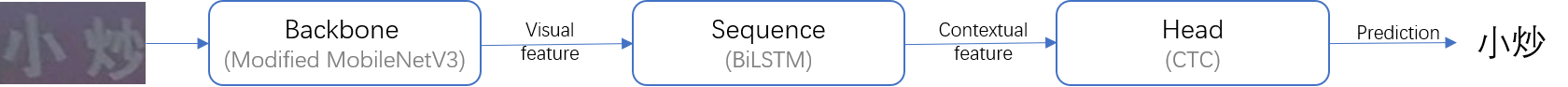}
    \caption{Architecture of the CRNN Algorithm}
    \label{fig:overview}
\end{figure}

 We use CRNN\cite{DBLP:journals/corr/ShiBY15} as our OCR framework, the architecture is shown in Figure \ref{fig:overview}.

\paragraph{Data Augmentation Design} We first studied the starter code provided in the competition and noticed the data augmentation provided baseline in PaddleOCR is not applicable for reasons below:
\begin{itemize}
  \item The provided augmentation setting does not align to data distribution as in our point of view. 
  \item The implementation is ill-designed: everything were coupled in a single function, which make it inconvenient for subsequent controlled experiment.
\end{itemize}

\par Based on the reasons above, we decide to redesign the data augmentation module. Note that we removed all data augmentations in over-fitting experiments to eliminate the inference from those augmentations. Input images are resized to 32 × 320 without any further augmentations.

\paragraph{Limit Character Category} Default category provided by PaddleOCR consists of 6623 characters. 
\begin{itemize}
  \item As shown in Table \ref{tab:amount}, since there are only 3974 unique characters in provided dataset, there is no need to keep characters that are not in the train set. This reduces the number of parameters and memory usage without any accuracy loss.
  \item We noticed that the space character contributes nothing during metric calculation in the competition, so we also removed the space characters from the Character Set.
\end{itemize}

\paragraph{Input/Output Setting} The size of input images is set to 32 × 320. And the max length of output sequence is set to 25 since the longest text in the data set is 24.

\paragraph{Model Architecture} Same as the default setting, we use MobileNetV3-Small\cite{DBLP:journals/corr/abs-1905-02244} with multipliers 0.5 as backbone, A two-layer BiLSTM with hidden size 48 as Neck(sequence encoder) and a fully connected layer as head.

\paragraph{Loss Function} We adopt CTC loss proposed by Graves \etal \cite{graves2006connectionist} as the loss function.

\paragraph{Training} We used Adam optimizer with cosine annealing with warm restarts:

\begin{itemize}
  \item Initial learning rate: $2e^{-3}$
  \item Total training epoch: $2000$
  \item Restart period: every $50$ epoch
\end{itemize}

\begin{table}[htb]
    \centering
    {
    \begin{tabular}{ccc}
\hline
            &Base Model &Large Model\\
\hline
Accuracy   &0.8096     &\textbf{1.0}\\
\hline\\
    \end{tabular}}
    \caption{Accuracy between different model capacity on Train Set}
    \label{tab:overfitting}
\end{table}

\par The settings above(called Base Model in Table \ref{tab:overfitting}) didn't over-fit the train set well. Conjecturing that here model capacity is the main shortcoming, we enlarged model capacity and did the experiment again:
\begin{itemize}
  \item Backbone: MobileNetV3-Large with multipliers $1.0$
  \item Neck: Increasing the hidden size of BiLSTM from $48$ to $256$
\end{itemize}
\par As shown in Table \ref{tab:overfitting}, the increase of model capacity brings training accuracy to $1.0$ which means it is more than enough for fitting the train set. We then continue to evaluate the accuracy on the val set. As shown in Table \ref{tab:baseline}, although the large model totally fitted the train set, the val accuracy seems acceptable. Since we plan to focus on accuracy first and shrink model size in the later stage, we choose this enlarged model as our baseline.

\begin{table}[htb]
    \centering
    {
    \begin{tabular}{ccc}
\hline
            &Base Model &Large Model\\
\hline
Accuracy   &0.5949     &\textbf{0.7494}\\
\hline\\
    \end{tabular}}
    \caption{Accuracy between different model capacity on Val Set}
    \label{tab:baseline}
\end{table}


\subsubsection{Data}

\par As data matters \textbf{a lot}, we start data-related experiments right after we got our baseline. At this stage, our goal is to further improve accuracy. Based on our observation and analysis, we conducted experiments on data augmentation and input image scale. 
\par For data augmentation, we focused on geometry transforms, noises, and color transforms. In natural scenes, text appears in various shapes and distorted patterns. In addition, the phenomenon of occlusion is also a big challenge for text recognition. The general approaches are adding data augmentation like Crop, Rotate, Cutout, TIA\cite{DBLP:journals/corr/abs-2003-06606}, etc. Noise, on the other hand, happens all the time in the real world, such as non-uniform illumination, blurring and electrical noises. We added different kinds of noise, blur to mimic them. Last but not least, to enrich color distribution in the dataset, we added some color-related transforms. Data augmentations we used are listed below:

\paragraph{Random Height Crop} We randomly crop the top or bottom of the image with respect to the Image size, the crop ratio is between 0 and 0.05.
\paragraph{Cutout} In order to alleviate the problem of text occlusion, We randomly replace 1 to 3 areas with random colors according to the image size, the ratio is between 0.05 and 0.1.
\paragraph{TIA} TIA\cite{DBLP:journals/corr/abs-2003-06606} is a data augmentation which is more effective and specific for training a robust recognizer. A set of fiducial points are initialized on the image, then these points are randomly be moved to generate new image with the geometric transformation, such as distortion, stretch and perspective. The number of initial fiducial points is randomly set between 3 and 6.
\paragraph{GaussNoise} We apply Gaussian noise with mean = 0 and standard deviation randomly between 0 and 10.
\paragraph{MotionBlur} We apply motion blur with kernel size between 3 and 7.
\paragraph{Order of augmentations} During data experiments, we adjusted the order of data augmentation:
\begin{itemize}
    \item Blurriness is influenced by the scale of the image. The effect of same blur kernel on different sized images differs a lot. If even a human eye can not recognize a blurred character, there is no way a model can recognize it. So we decide to put the MotionBlur after image resize to make sure same kernel is applied to images with same scale.
    \item The gaussian noise after blur looks very different from data distribution in the val set. Our experiment also proves that put GaussNoise before MotionBlur is a better choice.
\end{itemize}

\par Aside from data augmentations mentioned, we adjusted the input size based on the analysis shown in Figure \ref{fig:train_image_height}: the height of $21.6\%$ of the images is greater than 48 pixels. To reduce information loss in the process of resize, we increased the input scale:

\paragraph{Input Scale} increase the input scale from 32 × 320 to 48 × 480 (Being limited by our hardware, we do not further increase the input scale).

\par A summary of experiments in this stage can be found in Table \ref{tab:data}.
\begin{table}[htb]
    \centering
    {
    \begin{tabular}{cccccccc}
    \hline
    Baseline$^*$&\checkmark&\checkmark&\checkmark&\checkmark&\checkmark&\checkmark&\checkmark\\
    Random Height Crop&&\checkmark&\checkmark&\checkmark&\checkmark&\checkmark&\checkmark\\
    Cutout&&&\checkmark&\checkmark&\checkmark&\checkmark&\checkmark\\
    TIA&&&&\checkmark&\checkmark&\checkmark&\checkmark\\
    GaussNoise&&&&&\checkmark&\checkmark&\checkmark\\
    MotionBlur&&&&&&\checkmark&\checkmark\\
    Input Scale&&&&&&&\checkmark\\
    \hline
    Accuracy&0.7494&0.7531&0.7588&0.7654&0.7679&0.7689&0.8039\\
    \hline
    \\
    \end{tabular}}\\
    \footnotesize{$^*$ Baseline is best model(Large Model) last Stage}\\
    \caption{Accuracy of Experiments on Data}
    \label{tab:data}

\end{table}

\subsubsection{Model}
\par This stage can be divide into two parts: higher accuracy, smaller model.
\par We noticed that in the backbone, the image is only down-sampled at the beginning and the end horizontally. Since, as shown in Figure \ref{fig:train_image_scale}, images are likely to be wider instead of taller, we speculate that the horizontal receptive field is not enough for this task. So we adjusted the receptive field:

\paragraph{Larger Receptive Field} In the first layer of the C2 stage of the backbone, we modified the stride of the down-sampling feature map from (2,1) to (2,2), which increases the receptive field horizontally.

\par Also, as the model is able to $100\%$ fit the train set in early experiments, we conjecture that our model is still suffering from over-fitting to some degree although a lot of augmentations were introduced. 

\paragraph{Back to Lightweight Model} Being required to submit a model less than 10M, we choose lowering the amount of parameters as our first attempt to reduce over-fitting. We decrease the multipliers of MobileNetV3-Large from 1.0 to 0.6 and cut down the hidden size of BiLSTM from 256 to 64. The number of parameters is reduced to 9.5M, which meets the requirements of the competition. Unexpectedly, as shown in Table \ref{tab:anti-overfitting}, we found that decreasing the multipliers aggravated over-fitting (val-accuracy dropped significantly while train-accuracy maintained roughly the same). We speculate that the reason behind this is model-wise double decent phenomena as explained in \cite{nakkiran2019deep} (By reducing parameters, our model are approaching 'Interpolation Threshold' from 'Modern Regime').

\begin{table}[htb]
\centering
{
\begin{tabular}{ccc}    
\hline
 &Large Model &Light Model\\
\hline
Train   &0.9992   &0.9957\\
\hline
Val     &0.8087   &0.7874\\
\hline\\
\end{tabular}}
\caption{Accuracy between Large and Light Model}
\label{tab:anti-overfitting}
\end{table}

\paragraph{Spatial Dropout} We applied spatial dropout \cite{tompson2015efficient} in all layers on C5 stage on backbone with $keep\_prob = 0.9$.

\paragraph{Weight Decay} We increased weight decay of the fully connected layer form $1e^{-5}$ to $4e^{-5}$.

\par A summary of results in this stage can be found in Table \ref{tab:model}.
\begin{table}[htb]
    \centering
    {
    \begin{tabular}{cccccc}
    \hline
    Fusion$^*$&\checkmark&\checkmark&\checkmark&\checkmark&\checkmark\\
    Larger Receptive Field&&\checkmark&\checkmark&\checkmark&\checkmark\\
    Light Model&&&\checkmark&\checkmark&\checkmark\\
    Spatial Dropout&&&&\checkmark&\checkmark\\
    Weight Decay&&&&&\checkmark\\
    \hline
    Accuracy&0.8039&0.8087&0.7847&0.8045&0.8112\\
    \hline
    \\
    \end{tabular}}\\
    \footnotesize{$^*$ Fusion is best model in Data Experiments}\\
    \caption{Accuracy of Experiments on Model}
    \label{tab:model}
\end{table}

\section{Full Dataset Training}
Generally we would like to do a cross-validation before using full data. However, due to limited time and machine, we trained the model on full dataset and submitted the checkpoint from the latest epoch in the end.

\section{Things We Tried But Didn't Work}
\paragraph{ColorJitter} Randomly changes the brightness, contrast, saturation, and hue of an image with default parameter setting from Albumentation\cite{info11020125}.
\paragraph{Pixel Reverse} Randomly reverse all the pixel of image.
\paragraph{Random Rotate} Rotate the image with a random degree between $(-10^{\circ}, 10^{\circ})$, and then resize the image to keep all pixels are still in the region of image.
\paragraph{Multi-Scale Feature Fusion} Concatenate the feature map for C3, C4 and C5 stage to merge low-level and high-level semantics.
\paragraph{Multi-Scale Training} Randomly select one from the following three scales ([32, 320], [48, 480] and [64, 640]) as the input scale during training, and replace the last pooling layer of the backbone to AdaptivePool to ensure that the size of the feature map is consistent. We switched scale every 8 iters during training.
\paragraph{Test Time Augmentation(TTA)} During inference stage we tried to use TTA on the multi-scale-trained model. We get predictions from the three scales mentioned above, and select the prediction with highest score as the final result. We also tried to average the predicted probabilities of the three before decoding, however these two methods did not bring any improvement.

\paragraph{Beam Search} In decode module we use Greedy Search algorithm which selects one best candidate as an input sequence for each time step. However, choosing just one best candidate might be suitable for the current time step, but when we construct the full sentence, it may be a sub-optimal choice. 
\par Based on this consideration, we tried to use Beam Search algorithm which selects multiple alternatives for an input sequence at each time step based on conditional probability. We tried different beam width, but none of them works.

\section{Conclusion}
Conclusively, we adopt CTC-based architecture, input scale of 48 × 480, series of data augmentation and anti-over-fitting methods, achieved 81.7\% on TestB data set in Ultra Light OCR Competition. We believe with more resources and time, the score could be further improved.
\bibliographystyle{unsrt}
\bibliography{template}
\end{document}